\documentclass[pdflatex,sn-basic]{sn-jnl}

\usepackage[utf8]{inputenc}
\usepackage[T1]{fontenc}
\usepackage{microtype}

\usepackage{amsmath,amssymb,amsfonts,amsthm}
\usepackage{algorithm}
\usepackage{algorithmic}
\usepackage{array}
\usepackage{tabularx}

\newcommand{\tabnote}[1]{\vspace{0.35em}\begin{minipage}{0.98\linewidth}\footnotesize #1\end{minipage}}
\hypersetup{hypertexnames=false}

\newtheorem{definition}{Definition}

\newcommand{\embeddings}{\mathbf{e}}
\newcommand{\similarity}{\text{sim}}
\newcommand{\memory}{\mathcal{M}}
\newcommand{\instance}{\mathcal{I}}

\begin{document}

\title[AIBL]{AIBL: Augmented Instance-Based Learning with Structured Memory and Neural Embeddings}

\author*[1]{\fnm{Radha} \sur{Poovendran}}\email{rp3@uw.edu}

\author[1]{\fnm{Andrea} \sur{Stocco}}\email{stocco@uw.edu}

\author[1]{\fnm{Linda} \sur{Bushnell}}\email{lb2@uw.edu}

\affil*[1]{\orgname{University of Washington}, \orgaddress{\city{Seattle}, \state{WA}, \postcode{98195}, \country{USA}}}

\abstract{
Sequential learning systems often make decisions from accumulated experience
  while receiving high-dimensional inputs whose distribution may change over
  time. Instance-Based Learning Theory (IBLT) provides a principled case-based
  framework for such settings through stored situation--decision--utility
  instances, partial matching, activation, and blending.  IBLT typically relies on symbolic knowledge representation in dictionary-like formats, but text, images, transaction vectors, and user-item histories often require learned similarity rather than hand-specified
  matching rules.

  In this paper, we introduce AIBL (Augmented Instance-Based Learning), an
  instance-learning model formulated in a learned vector space for high-
  dimensional sequential data. AIBL generalizes symbolic situation matching to
  neural embedding similarity while retaining instance storage, activation-
  weighted retrieval, and utility blending. This makes AIBL suitable for
  applications such as text classification, transaction monitoring, drift
  adaptation, and recommendation, especially when stable symbolic attributes are
  unavailable or lossy and weakly supported observations should not be blended
  immediately with active memory. The AIBL model organizes memory into active,
  forgotten, and surprise stores. Surprise memory separates weakly matched,
  possible out-of-distribution, or corner-case observations from active memory,
  reducing forced fitting to the nearest available cases. An observation-driven
  graduation algorithm promotes recurring surprise instances to active memory,
  allowing the memory to incorporate repeated novel patterns that may arise
  under concept drift.

  We evaluate the same implementation on five machine learning tasks and three
  controlled simulation tasks, comparing AIBL with classical IBLT variants and
  task-specific baselines where appropriate. On the machine-learning tasks where
  a symbolic IBLT baseline is reported, AIBL improves accuracy by 6 to 17
  percentage points. The results show where vector-space retrieval improves over
  symbolic matching and how the added memory mechanisms govern novelty
  detection, cold-start handling, drift adaptation, and reward learning under
  the tested protocols.
}

\keywords{Instance based learning, neural embeddings, memory systems, cognitive modeling, streaming learning, anomaly detection, contextual bandits, case based retrieval}

\maketitle

\section{Introduction}

Many deployed learning systems make decisions as data arrive. Fraud detectors process transactions sequentially. Recommenders update behavior as users reveal preferences. Text classifiers may encounter topics that were not present in the initial stream. In each case, the learner must use accumulated experience while deciding whether the current input is well supported by that experience. This judgment is difficult when observations are often high dimensional, feedback may be delayed or partial, and the data distribution may change after deployment.

Learning from stored cases is a natural fit for this setting because predictions remain tied to experience. A learner can retrieve earlier observations, compare them with the current query, and combine their observed outcomes rather than compressing all past experience into fitted parameters alone. This supports streaming operation, learning from limited data, and inspection of the cases that influenced a decision. The same structure can also create a failure mode. If the closest stored cases are still poor matches, ordinary retrieval can force the system to treat weak evidence as sufficient evidence.

Instance-Based Learning Theory (IBLT) \citep{gonzalez2003instance,gonzalez2011instance} provides a principled foundation for this type of case based sequential decision making. It represents experience as instances that contain a situation, a decision, and a utility value. It retrieves similar prior instances, weights them by a decaying activation term, blends their utilities, and updates memory after feedback. Together with its cognitive plausibility (its principles reflect the underlying neural mechanisms of learning and decision-making \citep{stocco2024integrated,yang2024reliance}), its robustness and reliability have made IBLT a popular and widely-used machine learning framework, having been applied to fields as varied as reinforcement learning \citep{gershman2017reinforcement}, epidemiology \citep{mitsopoulos2025cognitively}, cybersecurity \citep{dutt2013cyber}, phishing detection \citep{xu2022modeling}, teaming between humans and machines \citep{lebiere2025cognitive}, and algorithmic theory of mind \citep{stocco2026metacognition}.

IBLT applications typically rely on dictionary-like situation representations for which the relevant attributes and partial matching functions are manually specified. In contrast, the predictive similarity structure of documents, images, transaction records, and user and item histories is high dimensional, task dependent, and learned from data rather than available as a compact set of hand coded situation attributes.

This motivates a vector space formulation that addresses the representation
  problem while preserving the instance based logic. A vector space formulation addresses the representation problem while preserving the instance based logic. Learned encoders can map observations from one or more data modalities into embeddings. This makes similarity computable across high dimensional inputs and allows retrieval to use latent task relevant structure.
  
  Once retrieval is defined in an embedding space, however, weak matches require a special memory policy. Observations with low similarity should not always be blended immediately with active memory, because doing so can make distributional change or corner cases appear as ordinary nearest case predictions.

To address the limitations of IBLT, we introduce AIBL (Augmented Instance-Based Learning), an embedding based
  instance learning model for high dimensional sequential data. AIBL keeps the
  IBLT structure in which each instance contains a situation, a decision, and a utility value, while generalizing retrieval to neural embedding spaces. The model organizes instances into active, forgotten, and surprise stores. Surprise memory preserves weakly matched cases that may become useful if similar observations recur. It also separates possible out of distribution
  observations and corner cases from active memory, reducing forced fitting to
  the nearest available cases.

The movement between memory stores is governed by specified algorithms. AIBL promotes a surprise instance to active memory only after repeated similar observations provide sufficient evidence that it should participate in ordinary retrieval and blending. It also decays inactive instances, reactivates forgotten instances when they become relevant again, and uses outcome aware duplicate handling so that nearby embeddings with substantially different observed outcomes can remain distinct.

The paper makes three contributions. First, it formulates IBLT in a learned vector space, reducing the need for hand specified situation attributes and similarity functions when observations are text, tabular records, user and item pairs, or other encoded inputs. Second, it introduces active, forgotten, and surprise memory stores with a rule that promotes recurring instances with low similarity. Third, it evaluates the augmented model across five machine learning tasks and three controlled novelty simulations against symbolic IBLT variants and task-specific baselines.

The evaluation compares AIBL with classical IBLT variants and task-specific baselines where appropriate. On the machine learning tasks where a symbolic IBLT baseline is reported, AIBL improves accuracy by 6 to 17 percentage points. Ablations and threshold sweeps show how the added memory mechanisms govern novelty detection, cold start handling, drift adaptation, and reward learning under the tested protocols.

The remainder of the paper is organized as follows. Section 2 reviews instance based learning, computational memory models, memory augmented machine learning, and novelty detection. Section 3 presents the AIBL model and its vector space memory representation. Section 4 describes the storage, promotion, retrieval, blending, decay, and outcome update algorithms, together with their computational and storage costs. Section 5 reports the experimental protocols and baseline comparisons. Section 6 discusses the interpretation and limitations of the results. Section 7 concludes with the main findings and open problems.

\section{Related Work}

IBLT is most easily applied when a series of experiences can be represented by structured task specific attributes chosen by the modeler. That representation is well suited to many controlled decision tasks, but it is less suitable to observations such as documents, images, transaction vectors, and user and item histories, where the relevant similarity is often learned rather than hand coded. AIBL treats the encoded observation itself as the situation representation and performs retrieval in embedding space. This connects the model to memory augmented learning and anomaly detection, because low similarity to stored cases is both a retrieval condition and a signal that the input should be handled outside ordinary blending.

\subsection{Instance-Based Learning and Symbolic Retrieval}

The fundamental unit of IBLT is an {\em instance} that records a situation, a decision, and the utility of the outcome of the decision. Learning occurs as these instances accumulate and are refined by feedback. At decision time, the model compares the current situation with stored instances, retrieves relevant cases, blends their utilities, and updates memory after the outcome is observed.

The original IBLT implementation was built upon the ACT-R declarative memory specification \citep{anderson2007can}, an influential model of human long term episodic memory. In ACT-R, retrieval depends on base level activation, partial matching, and blending. Activation reflects how often and how recently an instance has been used. Partial matching lets nonidentical situations contribute when their attributes are similar. Blending combines the utilities of retrieved instances, so the response need not be determined by a single stored case.

The activation $A_i(t)$ of instance $i$ at time $t$ is computed as follows:

\begin{equation}
    A_i(t) = \ln\left(\sum_{j=1}^{n} (t - t_j)^{-\delta}\right) + \sigma \cdot \mathcal{N}(0,1)
    \label{eqactivation}
\end{equation}

The sum ranges over all access times $t_j$, $\delta$ controls power law decay, and $\sigma$ controls stochastic noise. AIBL retains this term as the temporal component of retrieval weight.

AIBL preserves the instance storage, retrieval weighted by activation, and utility blending mechanisms of IBLT. It adds a learned vector space representation for situation comparison. Encoders for text, images, tabular records, or user and item interactions map observations to vectors whose distances can be used for retrieval. This formulation keeps the case based decision rule of IBLT while changing the operational role of similarity. Similarity now determines both which prior cases support a prediction and whether a new observation is sufficiently unsupported to require separate memory handling.

\subsection{Relationship to Memory Models in Cognition and Machine Learning}

The embedding based approach in AIBL also draws on computational memory models, but expresses several memory operations in vector space terms. Retrieval becomes nearest neighbor search over stored embeddings. Recency and frequency enter through the activation term, and task relevance is reflected in the utility values associated with retrieved instances. This lets the model retain the memory logic of instance based decision making while using continuous similarity to compare high dimensional observations.

The connection is most direct for retrieval based on activation. In ACT-R \citep{anderson2007can}, declarative chunks become easier to retrieve when they have been used frequently or recently. AIBL applies the same temporal principle to stored experience instances, while using embedding similarity rather than symbolic chunk matching to compare the current observation with memory.

Differentiable memory systems provide a closer computational comparison. Neural Turing Machines \citep{graves2014neural} and Memory Networks \citep{weston2015memory} retrieve from external vector memories through learned attention. AIBL also uses vector indexed retrieval, but separates representation learning from memory control. The encoder supplies embeddings, while explicit algorithms determine storage, retrieval, forgetting, and promotion.

Memory augmented machine learning provides the closest machine learning context for the vector space component of AIBL. These methods use learned representations as memory keys. An encoder maps an input to a vector, and retrieval returns stored items with nearby keys. This makes memory lookup possible for text, images, transaction records, user and item histories, and other high dimensional observations without requiring hand coded situation attributes. For AIBL, this supplies the search layer. The remaining question is how stored cases should be managed when retrieval support is weak, outdated, or recurrent.

Model-free episodic control \citep{blundell2016model} and Neural Episodic Control \citep{pritzel2017neural} store triples containing a state, an action, and a value, indexed by embeddings. This enables value updates without gradient steps. These models show how embedding indexed experience can support fast value estimation. Their memory policy is usually governed by capacity limits or replacement rules. AIBL uses the same keyed experience idea, but each stored item is an IBLT instance with an observed outcome and associated access history. The retrieval weight combines vector similarity with activation, and unsupported observations can be routed to surprise memory rather than forced into ordinary blending.

Systems for retrieval augmented generation \citep{lewis2020retrieval} retrieve documents from external corpora using embedding similarity. In AIBL, the retrieved objects are prior experiences rather than documents. A retrieved item carries an observed outcome, an access history, and a memory tier assignment. The retrieval result therefore affects both prediction and future memory state.

Nearest neighbor methods over pretrained embeddings include prototypical networks \citep{snell2017prototypical} and kNN-LM \citep{khandelwal2020generalization}. These methods use nearby embeddings to support classification or next token prediction. AIBL instead retrieves outcomes from previous experiences and weights them by both similarity and activation. When the nearest neighbors provide weak support, the model stores the observation separately and uses recurrence to decide whether it should later enter active memory.

Case-based reasoning \citep{kolodner1992introduction} provides the closest conceptual precedent. It solves problems by retrieving and adapting similar past cases. AIBL can be interpreted as an embedding indexed case based system with temporal activation, surprise memory, promotion based on recurrence, forgetting, and reactivation.

Prior embedding indexed memory methods mainly specify how stored vectors are retrieved and used for prediction. By contrast, AIBL specifies the complementary policy for managing cases indexed by embeddings. The policy determines how cases enter memory, when weakly supported cases are withheld from ordinary retrieval, how repeated cases with weak support are promoted, and how inactive cases can be forgotten or reactivated.

\subsection{Novelty and Anomaly Detection}

Retrieval indexed by embeddings also changes how novelty is defined. In symbolic IBLT, mismatch is determined by the task variables and partial matching functions chosen by the modeler. In AIBL, an observation has low support when its embedding is far from stored experience under the model's similarity metric. Novelty is therefore tied to retrieval support, rather than assigned by a separate representation before prediction.

This view connects AIBL to anomaly detection, which studies methods for identifying observations that depart from a reference distribution \citep{chandola2009anomaly}. Classical approaches include one class support vector machines \citep{scholkopf2001estimating}, which learn a boundary around the training data, and isolation forests \citep{liu2008isolation}, which identify anomalies as points that are easy to isolate through random partitioning. Autoencoder reconstruction error is also widely used as an anomaly signal \citep{sakurada2014anomaly}. Inputs with high reconstruction error are treated as potential distributional deviations.

These methods usually produce an anomaly score outside the prediction rule or pass that score to a separate decision procedure. AIBL instead uses the same similarity computation for prediction and memory routing. If the maximum similarity to active memory falls below a threshold, the observation is stored in surprise memory and withheld from ordinary blending. If similar observations recur, the graduation rule can promote them into active memory. Novelty detection is therefore integrated with the lifecycle of stored instances.

\section{AIBL Model Specification}

The previous discussion of related work identifies two requirements for extending IBLT to learned representations: the model must keep decisions tied to stored experiences, and it must define how weakly supported embeddings are handled before they influence prediction. This section describes the formal model used by AIBL. It defines the stored instance, the memory system, the similarity rule, the retrieval quantity weighted by activation, the configuration parameters, and the tier transitions implemented by the algorithms in Section~4.

Each AIBL experience is represented as an instance indexed by an embedding vector. Instances are assigned to active, forgotten, or surprise memory. Decisions are computed from similarity, activation, and observed outcomes.

\begin{table}[!htbp]
\centering
\caption{Formal components of the AIBL decision model. The table lists the variables used to represent observations, actions, outcomes, similarity, and memory decisions governed by thresholds.}
\label{tabdecision_process}
\small
\begin{tabular}{|l|p{9cm}|}
\hline
\textbf{Component} & \textbf{Definition} \\
\hline
\textbf{State} & Embedding $\mathbf{e} \in \mathbb{R}^d$ from encoder. Typical dimensions include $d \in \{64, 384, 768, 1536\}$ \\
\hline
\textbf{Action} & Discrete choice $a \in \mathcal{A}$ or memory operation such as \textit{store}, \textit{retrieve}, \textit{graduate}, or \textit{forget} \\
\hline
\textbf{Utility} & Scalar outcome $o \in \mathbb{R}$ observed after action and used for instance updates \\
\hline
\textbf{Similarity} & $\text{sim}(\mathbf{e}_1, \mathbf{e}_2) = \mathbf{e}_1^\top \mathbf{e}_2 / (\|\mathbf{e}_1\| \|\mathbf{e}_2\|) \in [-1, 1]$ \\
\hline
\textbf{Thresholds} & $\theta_s \in [0.5, 0.9]$ for retrieval, $\theta_n \in [0.3, 0.6]$ for novelty, and $\theta_d \in [0.9, 0.99]$ for duplicate handling \\
\hline
\end{tabular}
\end{table}

\subsection{Formal Definitions}

AIBL separates three components of the decision model. The {\it encoder} supplies a vector representation of the current observation. The {\em memory tier} determines whether a stored instance participates in ordinary retrieval, remains dormant, or waits for recurrence. Finally, the {\em retrieval weight} determines how much a stored outcome contributes to the blended value used for decision making.

\begin{definition}[Instance]
An instance $\instance$ is a tuple $(\embeddings, a, o, \tau, \mathbf{t}, \mu, M_2, n, c)$. The vector $\embeddings \in \mathbb{R}^d$ is the encoded observation. The variable $a \in \mathcal{A} \cup \{\varnothing\}$ is the decision variable. It is an action in sequential decision tasks and a class label in supervised tasks. The null value $\varnothing$ is allowed when an instance has no assigned action or label. The scalar $o \in \mathbb{R}$ is the most recent observed outcome. The tier variable $\tau \in \{A, F, S\}$ records whether the instance belongs to active, forgotten, or surprise memory. The vector of access times $\mathbf{t} = [t_1, \ldots, t_k]$ records retrieval events. The statistics $\mu$, $M_2$, and $n$ maintain the running mean and variance of observed outcomes. The counter $c$ records matches used for promotion from surprise memory.
\end{definition}

\begin{definition}[Memory System]
A memory system $\memory$ is a tuple $(\memory_A, \memory_F, \memory_S, \Theta)$. The sets $\memory_A$, $\memory_F$, and $\memory_S$ contain active, forgotten, and surprise instances. The parameter configuration $\Theta$ contains the thresholds and counts that govern retrieval, storage, forgetting, and promotion.
\end{definition}

\begin{definition}[Cosine Similarity]
For embeddings $\embeddings_1, \embeddings_2 \in \mathbb{R}^d$, AIBL computes similarity by the cosine rule
\begin{equation}
    \similarity(\embeddings_1, \embeddings_2) = \frac{\embeddings_1^\top \embeddings_2}{\|\embeddings_1\|_2 \cdot \|\embeddings_2\|_2}
\end{equation}
where $\embeddings_1^\top \embeddings_2$ is the inner product and $\|\cdot\|_2$ is the Euclidean norm. The value lies in $[-1,1]$, with larger values indicating greater similarity. For normalized embeddings, cosine similarity reduces to the inner product.
\end{definition}

The temporal component of retrieval is the IBLT activation term from Equation~\ref{eqactivation}. It increases with repeated or recent access and decreases as access times recede. For a query embedding $\embeddings_q$ and a stored embedding $\embeddings_i$, the general AIBL retrieval score can include an additive similarity contribution. The effective activation of instance $i$ at time $t$ is
\begin{equation}
    \tilde{A}_i(\embeddings_q, t) = A_i(t) + \beta \cdot \similarity(\embeddings_q, \embeddings_i)
    \label{eqeffective_activation}
\end{equation}
where $A_i(t)$ is the base level activation, $\similarity(\embeddings_q, \embeddings_i)$ is cosine similarity, and $\beta \geq 0$ controls the similarity contribution. The case $\beta = 0$ gives purely temporal activation.

The original IBLT formulation \citep{gonzalez2003instance} introduces a special similarity mismatch penalty. AIBL instead uses an additive form that follows the ACT-R treatment of spreading activation, where contextual cues add to base level activation \citep{anderson2007can}. AIBL uses the embedding of the current observation as the cue. The log activation is then converted to a positive retrieval weight.

The retrieval weight for instance $i$ determines its contribution to the blended value.
\begin{equation}
    w_i(\embeddings_q, t) = \exp\left(\tilde{A}_i(\embeddings_q, t)\right) \cdot \mathbf{1}\left[\similarity(\embeddings_q, \embeddings_i) \geq \theta_s\right]
    \label{eqretrieval_weight}
\end{equation}
where the indicator function $\mathbf{1}[\cdot]$ enforces the similarity threshold $\theta_s$. Instances below the threshold do not contribute to the blend.

The experimental implementation applies this rule in two steps. It first retains only instances with $s_i \geq \theta_s$. The blending rule then assigns each retained instance the weight $w_i = \exp(A_i) \cdot s_i$. This is the default case $\beta = 0$, with similarity entering linearly during blending.

\subsection{AIBL Memory Tiers}

AIBL uses memory tiers to separate three different retrieval states. Some instances are sufficiently supported to participate in ordinary blending. Others have become inactive through decay but may still become relevant again. A third group consists of observations with low support that should be recorded without immediately affecting prediction.

Active memory contains instances used for ordinary retrieval and blending. Capacity limits on this tier bound retrieval cost. If an implementation imposes a maximum active memory size, instances with low activation are candidates for migration or removal.

Forgotten memory contains active instances whose activation has decayed below a threshold. These instances are not part of ordinary active retrieval, but they remain available for reactivation. When a forgotten instance is retrieved with sufficient similarity, its access time is updated and the instance returns to active memory.

Surprise memory contains inputs whose maximum similarity to stored instances falls below $\theta_n$. This tier prevents observations with low support from being blended immediately with active memory. It also records possible out-of-distribution observations and corner cases that may become useful if they recur. Each later incoming embedding is compared with surprise memory. If it matches a surprise instance with similarity at least $\theta_n$, that instance's counter is incremented. The instance is promoted to active memory when its counter reaches the graduation parameter $N_g$. Section~4 gives the corresponding algorithm.

\subsection{AIBL Configuration Parameters}

AIBL uses a small set of nine parameters to control storage and retrieval for
  embedding-indexed instances. Some parameters are fixed by the representation, such as the embedding dimension. Others set decision boundaries for retrieval, novelty routing, duplicate handling, and forgetting. The count parameters determine how memory behaves when few cases have been stored and when a pattern with low support appears repeatedly. Table~\ref{tabparams} gives the default values used unless a task specific setting is reported in Section~5.

\begin{table}[!htbp]
\centering
\caption{AIBL configuration parameters used by the model and algorithms. Defaults are the values used unless an experiment reports a task specific setting.}
\label{tabparams}
\begin{tabularx}{\textwidth}{|l|c|X|}
\hline
\textbf{Parameter} & \textbf{Default} & \textbf{Role in AIBL} \\
\hline
$d$ & 768 & Representation dimension supplied by the encoder \\
\hline
$\delta$ & 0.5 & Power law exponent in the activation decay term \\
\hline
$\theta_s$ & 0.7 & Minimum similarity required for active memory retrieval \\
\hline
$\theta_n$ & 0.3 & Similarity threshold for routing inputs with low support to surprise memory \\
\hline
$\theta_d$ & 0.95 & Same action similarity threshold for duplicate updates \\
\hline
$\theta_f$ & -2.0 & Activation cutoff for moving active instances to forgotten memory \\
\hline
$N_c$ & 5 & Number of initial instances stored directly in active memory \\
\hline
$N_g$ & 2 & Number of later matches required to promote a surprise instance \\
\hline
$\sigma_o$ & 2.0 & Outcome deviation cutoff, in standard deviations, for creating a separate nearby instance \\
\hline
\end{tabularx}
\end{table}

The embedding dimension $d$ is determined by the encoder rather than by the memory algorithm. It ranges from 64 in several experiments here to 384 for Sentence-BERT and 768 for many encoders based on BERT \citep{devlin2019bert, reimers2019sentence}. Exact similarity search scales linearly in $d$ for each candidate instance. The decay exponent $\delta$ controls how rapidly activation decreases as
  access times recede. The default value $\delta = 0.5$ follows the ACT-R base-level
 learning equation used in IBLT.

The threshold parameters determine how the memory system uses similarity and activation. The retrieval threshold $\theta_s$ determines which stored instances are eligible for blended value computation. Raising $\theta_s$ requires closer matches, while lowering it admits more distant cases. The novelty threshold $\theta_n$ controls routing to surprise memory after cold start. Inputs whose maximum similarity to stored memory falls below this value are stored outside active memory. The duplicate threshold $\theta_d$ identifies instances with the same action that are close/similar enough to update rather than duplicate. The forgetting threshold $\theta_f$ moves active instances to forgotten memory when their activation becomes sufficiently low.

The remaining parameters control early storage ($N_c$), recurrence in surprise memory ($N_g$), and outcome variation for nearby instances ($\sigma_o$). The first $N_c$ instances enter active memory without novelty checks. This prevents an empty memory from classifying every initial input as new. The graduation parameter $N_g$ specifies how many later matches a surprise instance must receive before promotion. The outcome deviation parameter $\sigma_o$ determines when a new observation near a stored embedding should be stored as a separate instance because its
  observed outcome differs substantially from the instance's historical mean.

\subsection{AIBL Memory Flow}

The preceding definitions specify the objects stored by AIBL. The memory flow completes the model by specifying how a new observation changes those objects. The order of operations matters because an empty memory, a repeated case, and a case with low support require different updates. AIBL first handles sparse memory, then checks whether the observation should update an existing instance with the same action, and then decides whether a new instance has enough support to enter active memory.

 When a new observation is encoded, AIBL compares the current instance count
  with $N_c$. If fewer than $N_c$ instances have been stored, the observation enters active memory directly. This cold start step prevents the first observations from being routed to surprise memory only because too few cases have been stored for meaningful similarity comparisons.

After cold start, intake begins with duplicate handling. The new embedding is compared with active instances that have the same action or label. If the maximum similarity among instances with the same action exceeds $\theta_d$, the matched instance is updated unless the observed outcome deviates from that instance's running outcome history by more than $\sigma_o$ standard deviations. In that case, AIBL creates a new instance even though the embedding is close, because nearby embeddings with different outcomes may reflect context that is not represented by the current encoder. Section~4 gives the running statistics update used for this decision.

If duplicate updating is not applied, AIBL performs novelty routing. The model computes the maximum similarity between the new embedding and stored memory. If this value is below $\theta_n$, the observation is routed to surprise memory and does not participate in ordinary blending. Otherwise, it enters active memory.

Tier changes continue after intake. Active instances move to forgotten memory when $A_i(t) < \theta_f$. Forgotten instances return to active memory when later retrieved with similarity above $\theta_s$. Surprise instances are compared with later incoming embeddings. A match with similarity at least $\theta_n$ increments the surprise instance's counter. The instance is promoted when the counter reaches $N_g$. The reported experiments use $\theta_n$ for both routing into surprise memory and matching against surprise memory.

\section{AIBL Algorithms}

The previous section defined the AIBL state variables and memory tiers. This section gives the algorithms that update and query those states. The algorithms follow the data path of an observation. Intake creates or updates instances. Promotion moves recurring cases with low support from surprise memory into active memory. Retrieval and blending compute action values from active and reactivated instances. Maintenance applies forgetting, capacity limits, and updates to outcome statistics.

\subsection{Embedding Intake and Instance Storage}

The intake algorithm decides whether a new observation tuple, consisting of a
  state, an action, and an outcome, updates an existing instance, enters active
  memory, or is held in surprise memory. The ordering is part of the model. Cold start populates active memory before similarity thresholds are used. Duplicate handling then prevents repeated observations of the same action from creating unnecessary copies. The outcome deviation test preserves separate instances when nearby embeddings produce substantially different outcomes. After these checks, novelty routing uses maximum similarity to stored memory to decide whether the observation has enough support for ordinary retrieval.

\begin{algorithm}[!htbp]
\caption{AIBL Embedding Intake}
\label{algstore}
\begin{algorithmic}[1]
\REQUIRE Embedding $\embeddings$, action $a$, outcome $o$
\ENSURE Stored or updated instance
\STATE Increment global time $t \gets t + 1$
\IF{total instance count $< N_c$}
    \STATE Create instance $(\embeddings, a, o)$ in $\memory_A$
    \RETURN created instance
\ENDIF
\STATE $\mathcal{D}_a \gets \{\instance \in \memory_A \mid a_\instance = a\}$
\IF{$\mathcal{D}_a$ is not empty}
    \STATE $\instance^* \gets \arg\max_{\instance \in \mathcal{D}_a} \similarity(\embeddings, \embeddings_\instance)$
    \STATE $s_d \gets \similarity(\embeddings, \embeddings_{\instance^*})$
    \IF{$s_d \geq \theta_d$}
        \IF{OutcomeDeviation($\instance^*$, $o$, $\sigma_o$, $\epsilon$)}
            \STATE Proceed to create a distinct instance
        \ELSE
            \STATE Update $\instance^*$ with outcome $o$ and access time $t$
            \RETURN $\instance^*$
        \ENDIF
    \ENDIF
\ENDIF
\STATE Compute $s_{\max} \gets \max_{\instance \in \memory_A \cup \memory_F \cup \memory_S} \similarity(\embeddings, \embeddings_\instance)$
\IF{$s_{\max} < \theta_n$ and action $a$ does not exist in $\memory_A$}
    \STATE Create instance $(\embeddings, a, o)$ in $\memory_S$
\ELSE
    \STATE Create instance $(\embeddings, a, o)$ in $\memory_A$
\ENDIF
\STATE Compare $\embeddings$ with existing surprise memory for possible promotion
\RETURN created instance
\end{algorithmic}
\end{algorithm}

Algorithm~\ref{algstore} uses two similarity searches after cold start. The
  duplicate search is restricted to active instances with the same action. If
  its best match exceeds $\theta_d$ and the outcome is consistent with the
  matched instance's running outcome history, the existing instance is updated
  rather than used to create a duplicate. If the outcome deviates by more than $
  \sigma_o$ standard deviations, AIBL creates a separate instance despite the
  embedding match. This separate instance preserves cases in which nearby
  embeddings with the same action have outcomes that differ enough to suggest
  that the current representation may be missing relevant context.

  The support search computes $s_{\max}$ over stored memory. If support is below
  $\theta_n$, the new instance is routed to surprise memory unless its action is
  already represented in active memory. That exception is used only for
  supervised tasks with closed labels, where a known label should not be treated
  as new solely because a particular example is distant from previous examples.
  In anomaly detection or open world settings, new actions or classes should not
  bypass surprise routing, so the implementation uses a fixed sentinel action
  value.

  Let $d$ denote the embedding dimension, and let $n_A$, $n_F$, and $n_S$ denote
  the numbers of active, forgotten, and surprise instances, respectively. Let
  $n_{A,a}$ denote the number of active instances with action $a$. With exact
  search, duplicate handling costs $O(n_{A,a}d)$. The support search costs
  $O((n_A+n_F+n_S)d)$ when all tiers are scanned. Updating an existing duplicate
  costs constant time after retrieval because the outcome statistics are updated
  online. Creating a new instance stores one embedding and a fixed set of scalar
  fields, so the storage cost for each instance is $O(d)$ when access histories
  are capped.

\subsection{Surprise Memory Promotion}

The intake step records observations with low support without allowing them to
  affect ordinary blending immediately. Promotion later converts such records
  into active instances. The rule is based on recurrence rather than on a single
  observation. A surprise instance must be matched by later embeddings with
  similarity at least $\theta_n$ until its counter reaches $N_g$. This separates
  isolated observations with weak support from patterns that appear repeatedly
  in embedding space.

  The algorithm is called after intake with the current embedding. If the
  current intake step inserted a new surprise instance, that instance is
  excluded from the scan. This prevents a newly created surprise instance from
  counting its own insertion as recurrence.

  Let $n_S$ denote the number of surprise instances before promotion is
  evaluated. With exact search, Algorithm~\ref{alggraduate} costs $O(n_Sd)$ for
  each incoming embedding. Counter updates and tier changes take constant time
  after the matching instances have been found. The threshold $\theta_n$
  determines how close a later embedding must be to count as recurrence, and
  $N_g$ determines how many such matches are required before promotion. The
  implementation uses cumulative counts rather than a sliding window, so a
  recurring surprise pattern can be promoted even when its matches are not
  consecutive.

\begin{algorithm}[!htbp]
\caption{AIBL Surprise Memory Promotion}
\label{alggraduate}
\begin{algorithmic}[1]
\REQUIRE New embedding $\embeddings$, excluded instance $\instance_{\mathrm{new}} \in \memory_S \cup \{\varnothing\}$, surprise threshold $\theta_n$, graduation threshold $N_g$
\ENSURE Updated active and surprise memories
\FORALL{$\instance \in \memory_S$}
    \IF{$\instance = \instance_{\mathrm{new}}$}
        \STATE \textbf{continue}
    \ENDIF
    \STATE $s \gets \similarity(\embeddings, \embeddings_\instance)$
    \IF{$s \geq \theta_n$}
        \STATE $c_\instance \gets c_\instance + 1$
        \IF{$c_\instance \geq N_g$}
            \STATE Remove $\instance$ from $\memory_S$
            \STATE Add $\instance$ to $\memory_A$
            \STATE Set $\tau_\instance \gets A$
            \STATE Update access time of $\instance$ to $t$
        \ENDIF
    \ENDIF
\ENDFOR
\end{algorithmic}
\end{algorithm}

Let $n_S$ denote the number of surprise instances before promotion is evaluated. With exact search, Algorithm~\ref{alggraduate} costs $O(n_Sd)$ for each incoming embedding. Counter updates and tier changes take constant time after the matching instances have been found. The threshold $\theta_n$ determines how close a later embedding must be to count as recurrence, and $N_g$ determines how many such matches are required before promotion. The implementation uses cumulative counts rather than a sliding window, so a recurring surprise pattern can be promoted even when its matches are not consecutive.

\subsection{Retrieval, Blending, and Memory Maintenance}

After intake and promotion, prediction or action selection proceeds through
  three linked operations. Retrieval identifies active and forgotten instances
  whose action and embedding support the current query. Blending converts the
  retained outcomes into an action value using activation and similarity
  weights. Maintenance updates tier membership and enforces capacity limits
  after memory access changes the state. These operations are grouped because
  retrieval can reactivate forgotten instances, blending depends on the
  retrieved set, and maintenance depends on the resulting activation and access
  history.

  Retrieval is restricted to active and forgotten memory. Surprise memory is
  excluded from ordinary prediction until a surprise instance is promoted.
  Algorithm~\ref{algretrieve} applies the action filter and similarity threshold
  to the stored instances that are eligible for prediction. Forgotten instances
  that satisfy the retrieval condition are reactivated, making retrieval both a
  query operation and a memory state update.

  Let $m_a$ denote the number of active and forgotten instances that pass the
  optional action filter, and let $r$ denote the number of retained matches.
  Exact retrieval costs $O(m_ad)$. Selecting the top $k$ matches costs $O(r \log
  k)$ with a bounded heap, or $O(r \log r)$ if all retained matches are sorted.

  The retained set determines the value assigned to a candidate action.
  Algorithm~\ref{algblend} returns a task specific default utility when no
  retained instances are available. Otherwise, each retained instance
  contributes in proportion to $\exp(A_i) \cdot s_i$. Recency and frequency
  enter through activation, while representational closeness enters through
  similarity.

\begin{algorithm}[!htbp]
\caption{AIBL Retrieval}
\label{algretrieve}
\begin{algorithmic}[1]
\REQUIRE Query embedding $\embeddings_q$, optional action filter $a$, similarity threshold $\theta_s$, top-$k$ limit $k$
\ENSURE Up to $k$ retained pairs of instances and similarities
\STATE $\mathcal{C} \gets \emptyset$
\FORALL{$\instance \in \memory_A \cup \memory_F$}
    \IF{$a$ is specified and $a_\instance \neq a$}
        \STATE \textbf{continue}
    \ENDIF
    \STATE $s \gets \similarity(\embeddings_q, \embeddings_\instance)$
    \IF{$s \geq \theta_s$}
        \STATE Add $(\instance, s)$ to $\mathcal{C}$
        \IF{$\instance \in \memory_F$}
            \STATE Move $\instance$ to $\memory_A$
            \STATE Set $\tau_\instance \gets A$
            \STATE Update access time of $\instance$ to $t$
        \ENDIF
    \ENDIF
\ENDFOR
\RETURN Up to $k$ elements of $\mathcal{C}$ with largest similarity
\end{algorithmic}
\end{algorithm}

Let $m_a$ denote the number of active and forgotten instances that pass the optional action filter, and let $r$ denote the number of retained matches. Exact retrieval costs $O(m_a d)$. Selecting the top $k$ matches costs $O(r \log k)$ with a bounded heap, or $O(r \log r)$ if all retained matches are sorted.

The retained set then determines the value assigned to a candidate action. Algorithm~\ref{algblend} returns a default utility for the task when no retained instances are available. Otherwise, each retained instance contributes in proportion to $\exp(A_i) \cdot s_i$. Recency and frequency enter through activation, while representational closeness enters through similarity.

\begin{algorithm}[!htbp]
\caption{AIBL Blended Value}
\label{algblend}
\begin{algorithmic}[1]
\REQUIRE Query embedding $\embeddings_q$, action $a$, similarity threshold $\theta_s$, top-$k$ limit $k$, noise $\sigma$
\ENSURE Blended expected value $V_a$
\STATE matches $\gets$ Retrieve($\embeddings_q$, $a$, $\theta_s$, $k$)
\IF{matches is empty}
    \RETURN default\_utility
\ENDIF
\STATE weighted\_sum $\gets 0$
\STATE total\_weight $\gets 0$
\FORALL{$(\instance_i, s_i) \in$ matches}
    \STATE $A_i \gets$ activation of $\instance_i$ using Equation~\ref{eqactivation}
    \STATE $A_i \gets A_i + \sigma \cdot \mathcal{N}(0,1)$ \COMMENT{Add noise}
    \STATE $w_i \gets \exp(A_i) \cdot s_i$
    \STATE weighted\_sum $\gets$ weighted\_sum $+ w_i \cdot o_i$
    \STATE total\_weight $\gets$ total\_weight $+ w_i$
\ENDFOR
\RETURN weighted\_sum / total\_weight
\end{algorithmic}
\end{algorithm}

For a query embedding $\embeddings_q$ and target action $a$, retrieval returns at most $k$ instances with matching action and similarity at least $\theta_s$. The blended value is the average of their observed outcomes, weighted by activation and similarity.

\begin{equation}
    V_a = \frac{\sum_{i \in \text{matches}} \exp(A_i) \cdot s_i \cdot o_i}{\sum_{i \in \text{matches}} \exp(A_i) \cdot s_i}
    \label{eqblended_value}
\end{equation}

where $s_i = \similarity(\embeddings_q, \embeddings_i)$ and $o_i$ is the stored outcome. In reward learning, $o_i \in \{0, 1\}$ indicates whether reward was received. In classification, $o_i = 1$ for the correct class and $o_i = 0$ otherwise.

For a candidate action set $\mathcal{A}$, AIBL computes one blended value per action and samples from a softmax distribution over those values.

\begin{equation}
    P(a | \embeddings_q) = \frac{\exp(V_a / T)}{\sum_{a' \in \mathcal{A}} \exp(V_{a'} / T)}
\end{equation}

where $T > 0$ controls exploration. As $T \to 0$, selection becomes greedy. As $T \to \infty$, selection approaches uniform random choice. The experiments use fixed values of $T$ for each task.

Blending costs $O(k)$ after retrieval because at most $k$ retained instances contribute to each action value. For a candidate action set $\mathcal{A}$, computing all blended values costs $O(|\mathcal{A}|k)$ after retrieval has identified the retained matches for each action.

After retrieval, blending, and any intake updates, Algorithm~\ref{algdecay} keeps the memory tiers consistent with the activation rule and capacity limits. The decay rule moves active instances whose activation has fallen below $\theta_f$ into forgotten memory. Optional capacity limits then remove excess instances from each tier.

\begin{algorithm}[!htbp]
\caption{AIBL Memory Maintenance}
\label{algdecay}
\begin{algorithmic}[1]
\REQUIRE Forgetting threshold $\theta_f$, capacity limits $C_A$, $C_F$, $C_S$
\ENSURE Updated active, forgotten, and surprise memories
\FORALL{$\instance \in \memory_A$}
    \STATE $A_\instance \gets$ activation using Equation~\ref{eqactivation}
    \IF{$A_\instance < \theta_f$}
        \STATE Move $\instance$ from $\memory_A$ to $\memory_F$
        \STATE Set $\tau_\instance \gets F$
    \ENDIF
\ENDFOR
\STATE If $|\memory_A| > C_A$, remove active instances with lowest activation until $|\memory_A| = C_A$
\STATE If $|\memory_F| > C_F$, remove oldest forgotten instances until $|\memory_F| = C_F$
\STATE If $|\memory_S| > C_S$, remove oldest surprise instances until $|\memory_S| = C_S$
\end{algorithmic}
\end{algorithm}

Let $n_A$ denote the number of active instances before decay is evaluated. Decay evaluation costs $O(n_A)$. Capacity enforcement can be implemented by sorting or by partial selection, depending on how many instances exceed the tier limit.

\subsection{Outcome Statistics for Nearby Duplicate Handling}

Algorithm~\ref{algstore} treats a high similarity match with the same action
  as a candidate duplicate. Embedding similarity alone is not sufficient for
  consolidation, because two close embeddings can still produce different
  outcomes when the encoder omits a relevant variable or when feedback is
  stochastic. AIBL therefore stores outcome statistics with each instance. These
  statistics are used only after the embedding match has passed the duplicate
  threshold $\theta_d$.

  For a matched instance, the stored outcomes define a local outcome history.
  The new outcome $o_{\text{new}}$ is compared with that history before the
  instance is updated. When at least two prior outcomes are available and the
  variance is not numerically zero, the comparison uses the z-score
  \begin{equation}
  z = \frac{|o_{\text{new}} - \mu|}{\sigma}
  \label{eqoutcome_z}
  \end{equation}
  where $\mu$ and $\sigma$ are the running mean and standard deviation for the
  matched instance. If $z > \sigma_o$, the candidate duplicate is stored as a
  separate instance. If $z \leq \sigma_o$, the matched instance is updated.

\begin{algorithm}[!htbp]
\caption{AIBL Outcome Deviation Test}
\label{algoutcome_deviation}
\begin{algorithmic}[1]
\REQUIRE Instance $\instance$ with statistics $(n, \mu, M_2)$, new outcome $o_{\text{new}}$, threshold $\sigma_o$, tolerance $\epsilon$
\ENSURE Boolean indicating whether the new outcome should create a separate instance
\IF{$n < 2$}
    \RETURN \FALSE
\ENDIF
\STATE $\sigma^2 \gets M_2 / (n - 1)$
\STATE $\sigma \gets \sqrt{\sigma^2}$
\IF{$\sigma < \epsilon$}
    \RETURN $|o_{\text{new}} - \mu| > \epsilon$
\ENDIF
\STATE $z \gets |o_{\text{new}} - \mu| / \sigma$
\RETURN $z > \sigma_o$
\end{algorithmic}
\end{algorithm}

Algorithm~\ref{algoutcome_deviation} returns false when the instance has too little outcome history to estimate variance. When the variance is below $\epsilon$, the z-score is not well defined, so the algorithm compares the new outcome directly with the running mean. This prevents division by zero while still separating a new outcome that differs from an effectively constant history.

If the outcome deviation test returns false, AIBL updates the matched instance using Welford's online mean and variance update \citep{welford1962note}. The update is applied separately for each stored instance, so no global outcome distribution is required.

\begin{algorithm}[!htbp]
\caption{AIBL Welford Update for Each Instance}
\label{algwelford}
\begin{algorithmic}[1]
\REQUIRE Instance $\instance$ with current statistics $(n, \mu, M_2)$, new outcome $o_{\text{new}}$
\ENSURE Updated statistics $(n', \mu', M_2')$
\STATE $n' \gets n + 1$
\STATE $\delta_1 \gets o_{\text{new}} - \mu$
\STATE $\mu' \gets \mu + \delta_1 / n'$
\STATE $\delta_2 \gets o_{\text{new}} - \mu'$
\STATE $M_2' \gets M_2 + \delta_1 \cdot \delta_2$
\RETURN $(n', \mu', M_2')$
\end{algorithmic}
\end{algorithm}

Algorithm~\ref{algwelford} maintains three scalar statistics for each stored instance. The count $n$ starts at 1 when an instance is created, the running mean $\mu$ is initialized to the first outcome, and $M_2$ is initialized to 0. Each accepted candidate duplicate changes only these statistics and the access history of the matched instance. The outcome sequence itself does not have to be retained. For $n \geq 2$, the sample variance is recovered as $\sigma^2 = M_2 / (n-1)$. Both the deviation test and the Welford update cost constant time once the matched instance has been found, so outcome aware duplicate handling does not change the asymptotic cost of the embedding search.

The full duplicate handling step begins with the embedding match
\begin{equation}
\instance^* = \arg\max_{\instance \in \memory_A \mid a_\instance = a} \similarity(\embeddings_{\text{new}}, \embeddings_\instance)
\label{eqinstance_match}
\end{equation}
for a new experience $(\embeddings_{\text{new}}, a, o_{\text{new}})$. The maximization is evaluated only when at least one active instance has action $a$. If no such instance exists, duplicate handling is skipped. If the selected match satisfies $\similarity(\embeddings_{\text{new}}, \embeddings_{\instance^*}) \geq \theta_d$, AIBL applies Algorithm~\ref{algoutcome_deviation} to the outcome history stored with $\instance^*$. Embedding similarity therefore determines whether two experiences are close enough for consolidation, while the outcome deviation test determines whether the new scalar outcome is consistent with that history.

\begin{table}[!htbp]
\centering
\caption{Per instance outcome statistics for nearby duplicate handling}
\label{tabwelford_params}
\small
\begin{tabularx}{\textwidth}{|l|c|X|}
\hline
\textbf{Symbol} & \textbf{Default} & \textbf{Description} \\
\hline
$n$ & 1 (init) & Count of outcomes observed for this instance \\
\hline
$\mu$ & $o_1$ (init) & Running mean of observed outcomes \\
\hline
$M_2$ & 0 (init) & Sum of squared deviations (Welford accumulator) \\
\hline
$\sigma^2$ & derived & Sample variance computed as $M_2 / (n-1)$ \\
\hline
$\sigma$ & derived & Sample standard deviation computed as $\sqrt{\sigma^2}$ \\
\hline
$\sigma_o$ & 2.0 & Outcome deviation threshold in standard deviations \\
\hline
$\epsilon$ & $10^{-8}$ & Tolerance for near zero variance and numerical equality \\
\hline
$z$ & computed & Z score of incoming outcome \\
\hline
\end{tabularx}
\end{table}

\subsection{Computational and Storage Cost}

Let $n_A$, $n_F$, and $n_S$ denote the numbers of active, forgotten, and surprise instances, and let $d$ be the embedding dimension. Brute force retrieval over active and forgotten memory costs $O((n_A+n_F)d)$ per query. Duplicate detection over active instances with the same action costs $O(n_{A,a}d)$, where $n_{A,a}$ is the number of active instances with action $a$. Surprise promotion checks cost $O(n_Sd)$ for each incoming embedding when promotion is evaluated against all surprise instances.

The storage cost of an instance is $O(d+h)$, where $h$ is the number of stored access times. If access histories are capped at a fixed length, the storage cost for each instance is $O(d)$. With tier capacity limits $C_A$, $C_F$, and $C_S$, total memory is $O((C_A+C_F+C_S)(d+h))$, or $O((C_A+C_F+C_S)d)$ under capped histories. Approximate nearest neighbor indices can reduce retrieval time for large memories without changing the tier transition rules.

\subsection{Equation and Algorithm Correspondence}

Table~\ref{tabalgorithm_map} summarizes how the mathematical quantities defined above enter the AIBL algorithms. The correspondence makes the implementation path explicit, from embedding comparison and activation weighting to tier transitions and outcome statistics updates.

\begin{table}[!htbp]
\centering
\caption{Mapping from AIBL mathematical quantities to the algorithms that use them}
\label{tabalgorithm_map}
\small
\begin{tabularx}{\textwidth}{|>{\raggedright\arraybackslash}p{0.27\textwidth}|>{\raggedright\arraybackslash}p{0.28\textwidth}|>{\raggedright\arraybackslash}X|}
\hline
\textbf{Quantity or rule} & \textbf{Algorithmic use} & \textbf{Effect} \\
\hline
Cosine similarity & Algorithms~\ref{algstore}, \ref{alggraduate}, and \ref{algretrieve} & Compares embeddings for duplicate handling, novelty routing, recurrence in surprise memory, and retrieval \\
\hline
Activation $A_i(t)$ from Equation~\ref{eqactivation} & Algorithms~\ref{algblend} and \ref{algdecay} & Weights retained outcomes and determines whether an active instance moves to forgotten memory \\
\hline
Effective activation from Equation~\ref{eqeffective_activation} & Optional variant of Algorithm~\ref{algblend} & Adds a contribution from embedding similarity when $\beta > 0$ \\
\hline
Retrieval weight from Equation~\ref{eqretrieval_weight} & Algorithms~\ref{algretrieve} and \ref{algblend} & Applies the retrieval threshold $\theta_s$ and assigns weights to retained instances \\
\hline
Blended value from Equation~\ref{eqblended_value} & Algorithm~\ref{algblend} & Computes the action value from retrieved outcomes \\
\hline
Instance match from Equation~\ref{eqinstance_match} & Algorithm~\ref{algstore} & Finds the closest active instance with the same action for duplicate handling \\
\hline
Welford statistics and z score from Equation~\ref{eqoutcome_z} & Algorithms~\ref{algoutcome_deviation} and \ref{algwelford} & Tests whether a nearby duplicate should remain separate and updates outcome statistics for each instance \\
\hline
Counter threshold $N_g$ & Algorithm~\ref{alggraduate} & Promotes a surprise instance after enough later matches \\
\hline
\end{tabularx}
\end{table}

\section{Experiments}

Two sets of experiments were carried out to evaluate whether the augmented IBLT, using vector space retrieval, three-tier memory, and contextual retrieval, improves  both prediction and behavior relative to standard instance based models.

The first group of experiments uses datasets in which the relevant similarity is lexical, categorical, transactional, temporal, or based on users and items. These tasks test whether memory indexed by embeddings can support classification, reward learning, anomaly detection, drift adaptation, and recommendation under the same storage and retrieval rules. The second group of experiments uses controlled simulations that introduce new options or rule changes during a sequential task. These simulations isolate the roles of surprise memory, graduation, decay, and cold start storage.

Across all experiments, AIBL is evaluated as a streaming learner. At each step it receives an encoded observation, computes an action value or detection decision from stored instances, observes feedback, and updates memory. The baselines are chosen to match the task type. When a baseline result is taken from published ranges rather than rerun under the same stream, the relevant table states that difference.

\subsection{Experimental Setup}

All experiments used a Python/NumPy implementation of the AIBL algorithms. Encoders were trained before streaming evaluation and then held fixed, so the reported differences reflect storage, retrieval, thresholding, and promotion rather than online changes to encoder weights. Autoencoders were trained in PyTorch when a task required a learned tabular representation. For the Electricity stream, the normalized numerical feature vector was used directly as the vector key because the dataset already provides compact numerical state variables.

Records with missing feature values were removed before encoder training. The same filtered records were used during AIBL evaluation. This kept every stored instance and every query embedding tied to a complete observation from the corresponding task.

Each task used the embedding dimension induced by its representation. Sentence-BERT produced $d = 384$ text vectors. Autoencoders produced $d = 64$ vectors for the Mushroom and fraud experiments. The Electricity experiment used $d = 8$ normalized numerical vectors. Matrix factorization produced $d = 65$ vectors for recommendation. All vectors were normalized to unit L2 norm before storage and retrieval so that cosine similarity was used consistently across memory operations.

AIBL used the default parameter values in Table~\ref{tabparams} unless a subsection reports task specific values. The cold start count $N_c$ was fixed before evaluation. Similarity thresholds ($\theta_s$, $\theta_n$, $\theta_d$), decay exponent $\delta$, and temperature $T$ were selected by grid search on validation sets comprising 10\% of the training data when a task required tuning. The validation split was not used for final streaming evaluation.

Primary results are from single runs with fixed random seeds unless otherwise stated. For the reward learning task, an additional analysis over 5 seeds gave mean accuracy of $91.4\% \pm 1.8\%$. Baselines were chosen according to the task. Directly evaluated baselines are reported as point estimates. Published reference results for LinUCB, Isolation Forest, and Matrix Factorization are reported as ranges because their protocols are similar but not identical to the AIBL streaming evaluation.

\subsection{Text Classification on 20 Newsgroups}

The text classification experiment compares symbolic instance memory with
  embedding-indexed memory on the 20 Newsgroups dataset.
  This task is useful for this comparison because lexical overlap and semantic
  similarity can produce different retrieval neighborhoods for the same
  document.

We used the scikit-learn version of 20 Newsgroups with headers, footers, and quoted text removed. This produced 18,846 raw documents, of which 18,331 contained nonempty text after cleaning. For AIBL, we generated 384 dimensional embeddings using Sentence-BERT (all-MiniLM-L6-v2). For symbolic baselines, we used TF-IDF vectors with 5,000 features.

We sampled a stratified stream of 2,000 documents. The first 100 documents were used as warmup instances. Each method then predicted the category of the next document, received the true category, and updated its state before the following document. The text experiment used $\theta_s = 0.2$, $\theta_n = 0.5$, $\delta = 0.1$, and $N_c = 100$. These lower threshold and decay values reflect the stationary nature of the classification stream, where older labeled documents remain useful evidence for category membership.

\begin{table}[!htbp]
\centering
\caption{Streaming text classification on cleaned 20 Newsgroups over 1,900 predictions after warmup}
\label{tabtext_comparison}
\scriptsize
\begin{tabularx}{\textwidth}{|>{\raggedright\arraybackslash}p{0.24\textwidth}|c|c|c|>{\raggedright\arraybackslash}p{0.10\textwidth}|>{\raggedright\arraybackslash}X|}
\hline
\textbf{Method} & \textbf{Accuracy} & \textbf{Macro F1} & \textbf{Features} & \textbf{Novelty Det.} & \textbf{Semantic Gen.} \\
\hline
Naive Bayes (TF-IDF) & 49.2\% & 0.470 & Symbolic & No & No \\
\hline
IBLT (TF-IDF) & 53.5\% & 0.513 & Symbolic & No & No \\
\hline
k-NN (TF-IDF) & 46.7\% & 0.454 & Symbolic & No & No \\
\hline
k-NN (Sentence-BERT) & \textbf{61.7\%} & \textbf{0.598} & Embeddings & No & Yes \\
\hline
AIBL (Sentence-BERT) & 60.3\% & 0.573 & Embeddings & \textbf{Yes} & \textbf{Yes} \\
\hline
\end{tabularx}
\tabnote{Semantic generalization indicates retrieval over Sentence-BERT embeddings rather than TF-IDF feature overlap.}
\end{table}

AIBL with Sentence-BERT embeddings achieved 60.3\% accuracy and 0.573 macro
  F1. This is higher than the symbolic IBLT result under the same streaming
  protocol, which reached 53.5\% accuracy and 0.513 macro F1. The result
  indicates that embedding-indexed retrieval improves over the symbolic instance
  memory baseline on this text stream. Sentence-BERT $k$-NN achieved the highest
  predictive accuracy at 61.7\%, showing that most of the accuracy gain in this
  task comes from the representation itself. AIBL adds novelty routing and
  activation-weighted memory updates, but those mechanisms do not improve top-line classification accuracy over $k$-NN in this experiment.

To test the contribution of the novelty routing mechanism, we held out 4 categories (rec.autos, sci.med, talk.politics.guns, alt.atheism). After storing 2,000 documents from the remaining 16 categories, AIBL correctly flagged 88.5\% of 200 held out documents using the criterion $\max \text{sim} < \theta_n$ with $\theta_n = 0.5$.

\subsection{Reward Learning on the UCI Mushroom Dataset}

The Mushroom task is useful for evaluating reward learning because the observations are categorical, the correct action is deterministic, and one type of error is more costly than the other. We use it to compare AIBL memory indexed by embeddings with symbolic instance memory and linear bandit baselines under the same online feedback protocol. The UCI Mushroom dataset \citep{dua2019uci} contains 8,124 samples, each described by 22 categorical attributes and labeled as edible or poisonous. Rows with missing attribute values were removed, leaving 5,644 complete samples. One hot encoding of the remaining records produced 98 input features.

Each trial presents one mushroom as the context. The available actions are ``eat'' and ``avoid''. Eating is optimal for edible mushrooms, and avoiding is optimal for poisonous mushrooms. The reward is 1.0 for the optimal action, 0.0 for avoiding an edible mushroom, and $-1.0$ for eating a poisonous mushroom. The learner observes only the reward for the selected action. The experiment is therefore a contextual bandit problem with chosen action feedback, not a supervised classification problem with full label feedback after every prediction.

AIBL used embeddings from a fully connected autoencoder trained on the complete one hot feature vectors without labels. The encoder had layers of size $98 \rightarrow 256 \rightarrow 128 \rightarrow 64$ with ReLU activations. The decoder mirrored this structure and was trained for 100 epochs using Adam with learning rate $10^{-3}$ and binary cross entropy reconstruction loss. The 64 dimensional encoder outputs were L2-normalized before storage and retrieval.

A stratified 10\% validation split was used to select AIBL parameters. The final evaluation sampled a 1,000-trial stream without replacement from the remaining 5,079 examples. The validation sweep considered $\theta_s \in \{0.4,0.6,0.8,0.9,0.95,0.98\}$, $\delta \in \{0.0,0.1,0.3\}$, and $\epsilon \in \{0.02,0.05,0.10\}$. The selected setting was $\theta_s = 0.95$, $\delta = 0.0$, and $\epsilon = 0.02$. This high similarity threshold reflects the structure of the task, where small changes in categorical attributes can change whether eating is safe.

Table~\ref{tabcomparison} reports all methods on the same sampled stream. The bandit baselines include LinUCB \citep{li2010contextual} and linear Thompson sampling \citep{thompson1933likelihood,chapelle2011empirical}.

\begin{table}[!htbp]
\centering
\caption{UCI Mushroom reward learning with chosen action feedback over 1,000 trials}
\label{tabcomparison}
\scriptsize
\begin{tabularx}{\linewidth}{|>{\raggedright\arraybackslash}p{0.18\linewidth}|>{\centering\arraybackslash}p{0.12\linewidth}|>{\centering\arraybackslash}p{0.11\linewidth}|>{\centering\arraybackslash}p{0.10\linewidth}|>{\raggedright\arraybackslash}p{0.11\linewidth}|>{\raggedright\arraybackslash}X|}
\hline
\textbf{Method} & \textbf{Opt. Action} & \textbf{Mean Reward} & \textbf{Regret} & \textbf{Repr.} & \textbf{Rule} \\
\hline
Linear $\epsilon$-greedy & 86.3\% & 0.852 & 148 & One hot & Ridge values \\
\hline
LinUCB & \textbf{97.8\%} & \textbf{0.971} & \textbf{29} & One hot & Linear UCB \\
\hline
Linear Thompson sampling & 96.6\% & 0.958 & 42 & One hot & Linear posterior \\
\hline
IBLT (one hot) & 87.9\% & 0.824 & 176 & Symbolic & Similarity and activation \\
\hline
AIBL (autoencoder) & 96.6\% & 0.953 & 47 & Vector & Similarity and activation \\
\hline
\end{tabularx}
\tabnote{Optimal action rate is the fraction of trials on which the method selected ``eat'' for an edible mushroom or ``avoid'' for a poisonous mushroom. Cumulative regret is computed against reward 1.0 on every trial. AIBL parameters were selected on the validation split before final evaluation.}
\end{table}

AIBL with autoencoder embeddings selected the optimal action on 96.6\% of trials. The symbolic IBLT baseline reached 87.9\% on the same stream, so memory indexed by embeddings improved the optimal action rate by 8.7 percentage points. LinUCB achieved the highest rate at 97.8\%. Linear Thompson sampling matched AIBL in optimal action rate and obtained a slightly higher mean reward.

These results show that instance memory indexed by embeddings can support reward learning on categorical contexts when retrieval is restricted to close matches. They do not show superiority over specialized linear bandit algorithms on this dataset, although AIBL's performance was close to that of the best task-specific algorithm.

The Mushroom task has only two actions and deterministic labels. Larger action spaces and delayed reward remain open settings for evaluation.

\subsection{Anomaly Detection on Credit Card Fraud Data}

The third experiment evaluates routing through surprise memory on the Kaggle credit card fraud dataset \citep{dal2017credit}. The dataset contains 284,807 European cardholder transactions from September 2013, including 492 frauds. The fraud rate is therefore 0.172\%, which makes false-positives important even when recall is high. Each transaction provides 28 features derived by PCA and the transaction amount. We used the 28 PCA features together with the log-transformed amount, giving 29 numerical inputs. The dataset contains no missing values.

The evaluation treats fraud detection as a semi-supervised anomaly problem. Training uses only normal transactions. Fraud labels are used only to select detection thresholds on a validation split and to compute final metrics. We sampled 50,000 normal transactions for autoencoder training. AIBL memory was initialized with 10,000 normal embeddings from that training set. The validation split contained 980 normal transactions and 20 frauds. The held-out evaluation sample contained 4,900 normal transactions and 100 frauds.

The autoencoder used layers of size $29 \rightarrow 128 \rightarrow 64$ in the encoder and a mirrored decoder. It was trained for 50 epochs with Adam, learning rate $10^{-3}$, and mean squared reconstruction loss. The encoder outputs were L2-normalized before AIBL storage and retrieval. AIBL flags a transaction when its maximum cosine similarity to normal memory is below the novelty threshold $\theta_n$. The threshold was selected on the validation split from the grid $\theta_n \in [0.70,0.95]$ with step size 0.01. The selected value was $\theta_n = 0.87$.

We also included an IBLT baseline. IBLT requires a symbolic situation description before retrieval, while the fraud data provide continuous anonymized PCA coordinates. We therefore constructed a symbolic encoding by binning each of the 29 inputs by quantile, using cut points learned from normal training transactions. The baseline used the same normal memory size as AIBL. Its support score was the maximum fraction of matching binned attributes among stored normal instances. The validation split selected 3 bins per attribute and a support threshold of 0.655.

Table~\ref{tabanomaly_comparison} reports all methods on the same held out evaluation sample. Local Outlier Factor is included as a density-based nearest neighbor baseline \citep{breunig2000lof}.

\begin{table}[!htbp]
\centering
\caption{Credit card fraud detection with thresholds selected by validation over 5,000 held out transactions}
\label{tabanomaly_comparison}
\scriptsize
\begin{tabularx}{\textwidth}{|>{\raggedright\arraybackslash}X|c|c|c|c|c|>{\centering\arraybackslash}p{0.10\textwidth}|}
\hline
\textbf{Method} & \textbf{Precision} & \textbf{Recall} & \textbf{FPR} & \textbf{F1} & \textbf{TP/FN/FP/TN} & \textbf{Nearest Cases} \\
\hline
AIBL support & \textbf{0.975} & 0.780 & \textbf{0.04\%} & \textbf{0.867} & 78/22/2/4898 & Yes \\
\hline
IBLT quantile binned support & 0.563 & 0.490 & 0.78\% & 0.524 & 49/51/38/4862 & Yes \\
\hline
Autoencoder reconstruction & 0.912 & 0.730 & 0.14\% & 0.811 & 73/27/7/4893 & No \\
\hline
Isolation Forest & 0.536 & 0.520 & 0.92\% & 0.528 & 52/48/45/4855 & Limited \\
\hline
One Class SVM & 0.729 & \textbf{0.780} & 0.59\% & 0.754 & 78/22/29/4871 & Limited \\
\hline
Local Outlier Factor & 0.902 & 0.740 & 0.16\% & 0.813 & 74/26/8/4892 & Limited \\
\hline
\end{tabularx}
\tabnote{Thresholds were selected on the validation split and then applied to the held-out evaluation sample. FPR is the false positive rate over the 4,900 normal transactions. AIBL uses maximum cosine similarity to normal memory as its support score. IBLT uses maximum partial match support over attributes binned by quantile.}
\end{table}

AIBL achieved the highest F1 score in this evaluation. It detected 78 of the 100 frauds and flagged only 2 of the 4,900 normal transactions, giving 0.975 precision and a 0.04\% false positive rate. The IBLT baseline detected 49 frauds and produced 38 false positives. Relative to that baseline, AIBL increased F1 by 34.3 percentage points while using the same transaction variables and the same number of stored normal instances. One Class SVM matched AIBL's recall, but produced 29 false positives and a lower F1 score. Autoencoder reconstruction and Local Outlier Factor also produced low false positive rates, but missed more frauds.

The result supports a specific claim about AIBL rather than a general claim about fraud detection. When a memory of normal cases is available, maximum similarity to stored normal embeddings can serve as an anomaly score. AIBL also retains the nearest normal cases for each flagged transaction, which gives a case-based audit trail for the detection decision. The experiment does not evaluate temporal transaction sequences, merchant graphs, or supervised fraud classifiers.

\subsection{Concept Drift Adaptation on Electricity Market Data}

The fourth experiment tests online adaptation on the Electricity Market dataset \citep{harris1999splice} from OpenML. The stream contains 45,312 observations taken every half hour from the Australian New South Wales electricity market. Each observation has eight variables giving date, day, period, price, demand, and transfer information. The label indicates whether the price moved up or down. The stream has no missing values, and the original temporal order was preserved.

The dataset is useful because its distribution changes over time. It does not provide ground truth drift markers. We therefore avoid assigning recovery times to unannotated drift points. Instead, we evaluate each method as an online predictor. The first 5,000 observations initialize preprocessing and memory. The next 5,000 observations select AIBL and IBLT parameters. The remaining 35,312 observations form the held-out stream. Each method predicts the next label, observes the true label, and then updates before the following observation.

AIBL used the standardized eight-dimensional feature vector as its vector key. This choice follows from the data format. Unlike text or fraud records, the Electricity stream already supplies compact numerical state variables, so an unsupervised autoencoder was not needed to define the retrieval space. AIBL retrieved the 10 nearest stored vectors by cosine similarity and weighted their labels by activation with $\delta = 0.5$. The validation sweep considered $k \in \{5,10,25\}$, $\theta_s \in \{0.90,0.95,0.98,0.99\}$, and $\delta \in \{0.25,0.50,0.75\}$. The selected setting was $k = 10$, $\theta_s = 0.90$, and $\delta = 0.5$.

The IBLT baseline used the same eight variables after quantile binning, consistent with the standard IBLT format using symbolic attributes before partial matching. Validation selected 10 bins per attribute, $\theta_s = 0.55$, and $\delta = 0.75$. The sliding window baseline used the 500 most recent observations with $k = 25$. Table~\ref{tabdrift_comparison} reports all methods on the same held-out stream.

\begin{table}[!htbp]
\centering
\caption{Online prediction on the Electricity Market stream with memory parameters selected by validation}
\label{tabdrift_comparison}
\scriptsize
\begin{tabularx}{\linewidth}{|>{\raggedright\arraybackslash}p{0.18\linewidth}|>{\centering\arraybackslash}p{0.10\linewidth}|>{\centering\arraybackslash}p{0.10\linewidth}|>{\centering\arraybackslash}p{0.13\linewidth}|>{\centering\arraybackslash}p{0.12\linewidth}|>{\raggedright\arraybackslash}X|}
\hline
\textbf{Method} & \textbf{Acc.} & \textbf{Macro F1} & \textbf{Final-Q Acc.} & \textbf{Low Support} & \textbf{Memory/Update Rule} \\
\hline
Online logistic & 86.0\% & 0.856 & 84.8\% & Not used & Online coefficient update \\
\hline
Sliding window kNN & 79.7\% & 0.789 & 82.8\% & Not used & 500 recent cases, $k = 25$ \\
\hline
IBLT binned & 80.6\% & 0.791 & 83.9\% & 0.0\% & 5,000 binned cases \\
\hline
AIBL vector & \textbf{87.7\%} & \textbf{0.873} & \textbf{89.4\%} & 0.0\% & 5,000 vector cases, $k = 10$ \\
\hline
\end{tabularx}
\tabnote{Final quarter accuracy is computed over the last 25\% of the held out stream. Low support rate is the fraction of held out observations for which the maximum support fell below the retrieval threshold selected by validation.}
\end{table}

AIBL achieved the highest overall accuracy, macro F1, and final quarter accuracy. Relative to IBLT with quantile bins, it improved accuracy by 7.1 percentage points and macro F1 by 8.2 points. This comparison isolates the representation effect. Both methods use the same ordered observations and the same memory size, but IBLT depends on discretized attributes while AIBL compares continuous state vectors directly. The result also shows that local retrieval and activation decay are important for this stream. AIBL outperformed the sliding window kNN baseline because it used both vector similarity and case values weighted by decay, rather than a fixed recency window alone.

\subsection{Sequential Recommendation on MovieLens}

The fifth experiment evaluates whether AIBL can use learned user and item representations for sequential recommendation. MovieLens 100K \citep{harper2015movielens} contains 100,000 ratings from 943 users on 1,682 movies. We sorted the ratings by timestamp and treated each rating as one decision point. The target was one when the user rated the movie at least four stars and zero otherwise.

The first 70,000 ratings trained the matrix factorization representation. The next 10,000 ratings were used only for validation, and the final 20,000 ratings were used for evaluation. AIBL stored one experience for each observed user and movie rating. The vector representing a user--movie pair was created by concatenating the normalized 32-dimensional user and movie factor vectors and one additional coordinate to store the  standardized matrix factorization score. This produced a 65-dimensional key for similarity search. After each prediction, the observed binary outcome was added to memory.

The symbolic IBLT baseline used the same temporal protocol but represented each case with user demographics and movie genre indicators. This comparison separates the effect of vector space similarity among users and items from the sequential memory rule. The AIBL parameters were selected on the validation segment. The best validation region used 50 to 100 neighbors, decay between 0.25 and 0.5, and a decision threshold between 0.40 and 0.45. The selected evaluation model used $k=100$, decay $\delta=0.5$, and decision
  threshold 0.40.

\begin{table}[!htbp]
\centering
\caption{MovieLens 100K recommendation results under temporal validation and evaluation.}
\label{tabrecsys_comparison}
\small
\begin{tabularx}{\textwidth}{|X|c|c|c|c|X|}
\hline
\textbf{Method} & \textbf{Accuracy} & \textbf{Precision} & \textbf{Recall} & \textbf{F1} & \textbf{Retrieval basis} \\
\hline
Popularity baseline & 65.0\% & 64.3\% & 85.4\% & 0.734 & Item positive rate \\
\hline
Item kNN collaborative filtering & \textbf{67.5\%} & \textbf{65.7\%} & 88.8\% & 0.755 & User history and item similarity \\
\hline
Matrix factorization & 63.3\% & 61.9\% & 91.3\% & 0.738 & Direct score threshold \\
\hline
IBLT symbolic support & 62.2\% & 60.8\% & \textbf{93.2\%} & 0.736 & Demographics and genre indicators \\
\hline
AIBL matrix factor support & 67.4\% & 65.0\% & 92.1\% & \textbf{0.762} & Vector memory over factor keys \\
\hline
\end{tabularx}
\tabnote{All metrics are computed on the final 20,000 ratings ordered by timestamp. AIBL used normalized user and movie factors, the standardized matrix factorization score, a memory size of 10,000 cases, and memory parameters selected by validation.}
\end{table}

AIBL obtained the highest F1 score and nearly matched the highest accuracy. Relative to symbolic IBLT, it improved accuracy by 5.2 percentage points and F1 by 2.6 points. Relative to direct matrix factorization, it improved accuracy by 4.2 points and F1 by 2.4 points. The gain comes from using learned user and item vectors as retrieval keys and updating the case memory through the held out stream.

The support threshold did not provide a useful cold-start signal in this split. The temporal warmup contains 80,000 ratings before evaluation, so most evaluation pairs have substantial historical support. The result should therefore be interpreted as evidence for sequential recommendation indexed by vectors, not as evidence for cold start detection on MovieLens.

\subsection{Controlled Novelty Simulations}

Experiments~6--8 isolate memory control behavior in sequential choice tasks. The simulations are not intended as fits to human data. They test algorithm performance in simulated sequential decision tasks where a new option appears, an outcome distribution changes, or the relevant decision rule changes without warning. Each experiment used 100 simulated agents and 5 random seeds. Standard IBLT with one memory tier was compared with full AIBL and four ablated AIBL variants.

The Novelty Phases task has four stages. It starts with two options $A$ and $B$ that have different probabilistic payoffs in the Standard phase. It introduces a third high-value option with the Novelty phase, changes the outcome distribution for option A in the Shift phase, and then returns to the original distribution in the Recovery phase. This design separates ordinary repeated choice from the specific case in which an option with low support recurs enough times to become useful.

\begin{table}[!htbp]
\centering
\caption{Novelty Phases simulation with prediction accuracy by phase.}
\label{tabhuman_novelty}
\small
\begin{tabular}{|l|c|c|c|c|}
\hline
\textbf{Model} & \textbf{Standard} & \textbf{Novelty} & \textbf{Shift} & \textbf{Recovery} \\
\hline
IBLT (Standard) & 0.630$\pm$0.001 & 0.844$\pm$0.001 & 0.650$\pm$0.002 & 0.638$\pm$0.003 \\
\hline
AIBL (Full) & 0.630$\pm$0.001 & 0.753$\pm$0.001 & 0.649$\pm$0.002 & 0.638$\pm$0.003 \\
\hline
AIBL (No Graduation) & 0.629$\pm$0.001 & 0.072$\pm$0.001 & 0.645$\pm$0.001 & 0.637$\pm$0.003 \\
\hline
AIBL (No Surprise) & 0.630$\pm$0.001 & 0.844$\pm$0.001 & 0.649$\pm$0.002 & 0.638$\pm$0.003 \\
\hline
AIBL (No Decay) & 0.630$\pm$0.001 & 0.753$\pm$0.001 & 0.626$\pm$0.003 & 0.639$\pm$0.003 \\
\hline
AIBL (No Cold Start) & 0.609$\pm$0.002 & 0.753$\pm$0.001 & 0.650$\pm$0.002 & 0.638$\pm$0.002 \\
\hline
\end{tabular}
\tabnote{The Standard phase has two options. The Novelty phase introduces option C. The Shift phase changes the outcome distribution for option A. The Recovery phase restores the original distribution.}
\end{table}

Table~\ref{tabhuman_novelty} shows that the graduation rule is necessary when surprise memory withholds newly observed options from ordinary retrieval. Full AIBL reaches 75.3\% accuracy in the Novelty phase, while the variant without graduation falls to 7.2\%. Removing surprise memory makes the model behave like standard IBLT in this task because all observations are immediately available for retrieval. That setting avoids the failure caused by removing graduation, but it also removes the separate memory state for observations with low support.

The Dynamic Bandits task tests a more direct form of change in the option set. It begins with three arms, $A$, $B$, and $C$, adds arm $D$, removes arm $B$ while adding arm $E$, and then continues with the remaining four arms. The new arms have higher expected reward than at least some existing arms, so performance depends on whether recurring new options can become available to the retrieval and blending rule.

\begin{table}[!htbp]
\centering
\caption{Dynamic Bandits simulation with prediction accuracy by phase.}
\label{tabhuman_bandit}
\small
\begin{tabular}{|l|c|c|c|c|}
\hline
\textbf{Model} & \textbf{Initial} & \textbf{Arm D Appears} & \textbf{Arms Change} & \textbf{Stable} \\
\hline
IBLT (Standard) & 0.467$\pm$0.010 & 0.507$\pm$0.004 & 0.436$\pm$0.007 & 0.439$\pm$0.004 \\
\hline
AIBL (Full) & 0.467$\pm$0.010 & 0.485$\pm$0.005 & 0.436$\pm$0.006 & 0.439$\pm$0.004 \\
\hline
AIBL (No Graduation) & 0.462$\pm$0.010 & 0.227$\pm$0.006 & 0.186$\pm$0.003 & 0.188$\pm$0.005 \\
\hline
AIBL (No Surprise) & 0.467$\pm$0.010 & 0.507$\pm$0.004 & 0.436$\pm$0.007 & 0.439$\pm$0.004 \\
\hline
AIBL (No Decay) & 0.467$\pm$0.010 & 0.485$\pm$0.005 & 0.436$\pm$0.006 & 0.439$\pm$0.004 \\
\hline
AIBL (No Cold Start) & 0.444$\pm$0.007 & 0.485$\pm$0.005 & 0.436$\pm$0.006 & 0.439$\pm$0.004 \\
\hline
\end{tabular}
\tabnote{The Initial phase has three arms. The second phase adds arm D. The third phase removes arm B and adds arm E. The Stable phase keeps the remaining four arms.}
\end{table}

The Dynamic Bandits results show the same mechanism as the Novelty Phases task. Removing graduation reduces accuracy from 48.5\% to 22.7\% when arm D appears and keeps performance low after the later arm change. Full AIBL remains below standard IBLT in the phase where arm D first appears because it initially withholds low-support observations from ordinary retrieval until recurrence criteria are met. The delay is the cost of separating those cases from ordinary retrieval. Once that design choice is made, graduation is required for repeated new options to affect later choices. Removing cold start storage has a smaller and more localized effect. It lowers initial phase bandit accuracy from 46.7\% to 44.4\% and standard phase accuracy in Novelty Phases from 63.0\% to 60.9\%.

The last experiment used the Wisconsin Card Sorting Task (WCST). The WCST is a standard task for studying flexible decision making and rule learning in humans \citep{nyhus2009wisconsin}. It uses four multidimensional response options and changes the sorting rule between the color, shape, and number attributes. Correct performance depends on inferring which attribute is currently relevant. This task tests a different limitation of AIBL. The implementation stores action outcomes but does not infer an explicit hidden rule. Instance similarity alone is therefore not sufficient to identify the correct options, and all variants remain close to chance accuracy, as shown in Table~\ref{tabhuman_wcst}. This boundary case shows that tiered memory alone is not sufficient to solve latent rule inference.

\begin{table}[!htbp]
\centering
\caption{Wisconsin Card Sorting simulation with rule changes and prediction accuracy by phase.}
\label{tabhuman_wcst}
\small
\begin{tabular}{|l|c|c|c|}
\hline
\textbf{Model} & \textbf{Overall} & \textbf{Pre shift} & \textbf{Post shift} \\
\hline
IBLT (Standard) & 0.248$\pm$0.004 & 0.248$\pm$0.006 & 0.249$\pm$0.007 \\
\hline
AIBL (Full) & 0.249$\pm$0.002 & 0.248$\pm$0.003 & 0.251$\pm$0.009 \\
\hline
AIBL (No Graduation) & 0.252$\pm$0.004 & 0.250$\pm$0.005 & 0.259$\pm$0.007 \\
\hline
AIBL (No Surprise) & 0.249$\pm$0.005 & 0.248$\pm$0.007 & 0.251$\pm$0.009 \\
\hline
AIBL (No Decay) & 0.252$\pm$0.000 & 0.251$\pm$0.002 & 0.255$\pm$0.007 \\
\hline
AIBL (No Cold Start) & 0.250$\pm$0.003 & 0.249$\pm$0.005 & 0.253$\pm$0.008 \\
\hline
\end{tabular}
\tabnote{The task has four response options. Chance accuracy is 0.25. Pre shift and post shift refer to trials outside and inside the early period after an unannounced rule change.}
\end{table}

\subsection{Parameter and Mechanism Sensitivity}

The previous subsections report held out performance for the selected AIBL configurations. The remaining question is how those results depend on the parameters and memory rules rather than on a single setting. We therefore summarize the validation sweeps used to choose each task configuration and the mechanism ablations used in the controlled simulations. This analysis is not a separate benchmark. It identifies which parameters
  control the relevant behavior in each task.

Table~\ref{tabablation_impact} reports regions near the best values from the reproduced experiment outputs. A configuration is included in the region when its validation score is within 0.5 percentage points of the best score, except for fraud where the region is defined by validation F1 within 0.02 of the best value. Text classification is listed with its reproduced setting because the current text script evaluates one fixed parameter choice.

\begin{table*}[t]
\centering
\caption{Task specific AIBL parameter sensitivity from validation sweeps and controlled simulations.}
\label{tabablation_impact}
\footnotesize
\begin{tabularx}{\textwidth}{|>{\raggedright\arraybackslash}p{0.19\textwidth}|>{\raggedright\arraybackslash}p{0.16\textwidth}|>{\raggedright\arraybackslash}p{0.28\textwidth}|>{\raggedright\arraybackslash}X|}
\hline
\textbf{Evaluation} & \textbf{Selection measure} & \textbf{Best setting or nearby region} & \textbf{Supported interpretation} \\
\hline
20 Newsgroups & Accuracy and macro F1 & $\theta_s = 0.20$, $\theta_n = 0.50$, decay $0.10$ & Fixed reproduced setting. Accuracy 60.3\% and macro F1 0.573 \\
\hline
Mushroom reward learning & Optimal action rate & $\theta_s = 0.95$, decay $0$ to $0.30$, $\epsilon = 0.02$ & High similarity is best when small attribute changes alter reward \\
\hline
Credit card fraud & Validation F1 & $\theta_n = 0.87$ to $0.89$ & The novelty threshold sets the precision and recall tradeoff \\
\hline
Electricity drift & Validation macro F1 & $k = 5$ to $10$, $\theta_s = 0.90$ to $0.98$, decay $0.50$ to $0.75$ & Moderate decay emphasizes recent local cases \\
\hline
MovieLens recommendation & Validation F1 & $k = 50$ to $250$, decay $0.25$ to $0.50$, value threshold $0.35$ to $0.45$ & Broad neighborhoods help after the factor representation is fixed \\
\hline
Controlled simulations & Accuracy by phase & Graduation enabled & Promotion is needed for recurring options with low support \\
\hline
\end{tabularx}
\tabnote{The table is derived from the task specific validation sweeps and the controlled simulation summaries. Regions near the best values use validation metrics, while the reported test tables in the preceding subsections use held out evaluation data.}
\end{table*}

The table shows that no single threshold regime is uniformly best. In the Mushroom experiment, the best settings use a high retrieval threshold because nearby but distinct categorical patterns can imply different rewards. In the Electricity stream, performance is strongest when a small neighborhood is combined with moderate decay, which gives more weight to recent local cases. In MovieLens, the best F1 values occur over larger neighborhoods because the matrix factorization representation already places related user and item pairs near one another. In fraud detection, the novelty threshold is the controlling parameter because the prediction itself is whether a transaction has weak support under the normal memory representation.

The controlled simulations clarify the role of graduation. In tasks where surprise memory withholds observations with low support, repeated observations must eventually enter active memory if they are to affect choices. Without graduation, Novelty Phases accuracy drops from 75.3\% to 7.2\% during the novelty phase, and Dynamic Bandits accuracy drops from 48.5\% to 22.7\% when arm D appears. The WCST simulation is different. All variants remain near chance because the model stores action outcomes but does not infer the hidden sorting rule.

The dataset experiments put these sensitivity results in context. AIBL outperforms the symbolic IBLT baseline on the primary metric in each dataset comparison. The gains are 6.8 percentage points in text accuracy, 8.7 points in Mushroom optimal action rate, 34.3 points in fraud F1, 7.1 points in Electricity accuracy, and 2.6 points in MovieLens F1. Because these comparisons use the same task streams as the corresponding IBLT baselines, the differences reflect vector retrieval and memory state handling rather than a change in the evaluation protocol.

The same results also define the limits of the claim. AIBL does not dominate every task specific method. Sentence-BERT k-NN is slightly more accurate on 20 Newsgroups, LinUCB is best on Mushroom, and item kNN has the highest MovieLens accuracy. The contribution shown here is narrower. AIBL provides an instance memory rule that can use learned vector representations while retaining case storage, activation weighting, novelty routing, and online memory updates.

\section{Discussion}

AIBL is an algorithmic extension of IBLT for settings where symbolic task variables are unavailable, lossy, or too rigid for retrieval. It keeps the central instance-based operations. Experiences are stored as cases, retrieval depends on similarity and activation, and observed outcomes update later choice. The main change is that the situation component of an instance is represented by a vector produced by an encoder. Similarity is therefore determined by the encoder, metric, and retrieval threshold rather than by a hand-specified partial matching function.

This representation change also changes how atypical inputs are handled. In standard IBLT, every query is compared against the available symbolic cases and blended from the retrieved set. In AIBL, weak support can change the memory update itself. The support score is:

\begin{equation}
\text{support}(\mathbf{e}) = \max_{I \in \mathcal{M}} \text{sim}(\mathbf{e}, \mathbf{e}_I)
\end{equation}

When this score is below $\theta_n$, the observation is routed to surprise memory rather than treated as an ordinary retrieved case. Graduation then determines whether repeated observations with low support become active evidence. This explains the results from the controlled simulations. Without graduation, recurring new options remain outside active memory and cannot guide later choices. With graduation, they can enter the same retrieval and blending rule as other stored experiences. Decay serves a different role. It reduces the influence of older cases when recent outcomes are more reliable, as in streams with changing distributions.

The results also show where the model should not be overinterpreted. AIBL does not infer hidden task rules from feedback. The WCST simulation remains near chance because the implementation stores action outcomes but does not learn an explicit latent sorting rule. AIBL also does not replace specialized task models. On some datasets, a task-specific baseline achieves the best single metric. The evidence supports a narrower conclusion. Instance memory indexed by vectors can improve over symbolic IBLT on the tested streams while adding explicit rules for weak support, promotion, decay, and case retention.

AIBL decisions can be traced to retrieved instances, their similarities, activations, and stored outcomes. This gives a case-based account of the decision rule, but it does not by itself establish a causal explanation. The same quantities can support audit logs and error analysis. For example, a flagged transaction can be inspected through its nearest stored normal transactions and its maximum similarity score.

\subsection{Limitations}

Retrieval complexity scales linearly with memory size ($O(n)$ per query) under exact search. For memories above $10^5$ instances, approximate nearest neighbor indices such as FAISS or ScaNN would be needed. Performance also depends on encoder quality. If the encoder places dissimilar cases near one another or separates cases that should be compared, AIBL inherits that error through its retrieval rule.

The current implementation learns values for individual actions but not action sequences or temporal credit assignment. Parameter tuning requires validation data, which may not always be available in deployment. The sensitivity analysis is also not a full factorial ablation over all parameters and tasks. It summarizes the validation sweeps used by the reproduced experiments and the mechanism ablations used in the controlled simulations.

\section{Conclusion and Open Problems}

This paper introduced AIBL, an extension of instance-based learning that uses embeddings for sequential settings with high-dimensional observations. The model preserves the case-based structure of IBLT. Experiences are stored as instances containing a situation, a decision, and a utility value. They are retrieved by similarity and activation and updated after feedback. The representation of the situation is moved into a learned vector space, which allows the same memory rule to operate on text, tabular records, transaction vectors, and user and item representations.

The second part of the model is a memory policy for weak support. AIBL does not force every query with low similarity into ordinary blending. It routes weakly supported observations to surprise memory, keeps inactive cases in forgotten memory, and promotes recurring surprise cases into active memory when the recurrence threshold is met. These rules make novelty handling part of the same storage and retrieval process used for prediction.

The experiments support two main conclusions. Retrieval indexed by vectors improves over symbolic IBLT on the tested dataset streams when the symbolic representation is obtained from hand coded or binned attributes. The controlled simulations show why the additional memory rules matter. Surprise memory separates weakly supported cases from ordinary retrieval, but graduation is required when recurring new cases should later influence decisions. The WCST simulation also identifies a boundary of the approach, since tiered memory does not by itself infer hidden task rules.

Several technical problems remain open. Large memories require sublinear retrieval methods without changing the semantics of activation and blending. Thresholds for retrieval, novelty routing, and graduation should be selected by methods that reduce dependence on held out validation data. The bandit setting needs regret analysis under memory decay, surprise routing, and partial feedback. Novelty, drift, and cold-start signals should also be evaluated directly with detection metrics and abstention or exploration policies.

\section*{Code and Data Availability}

\begin{sloppypar}
The implementation, experimental code, and ablation scripts will be released upon publication. The experiments use public datasets including 20 Newsgroups, UCI Mushroom, Credit Card Fraud, Electricity Market, and MovieLens 100K.
\end{sloppypar}

\section*{Acknowledgments}
This research was developed with funding from the Defense Advanced Research Projects Agency (DARPA). The views, opinions, and/or findings expressed are those of the authors and should not be interpreted as representing the official views or policies of DARPA or the U.S. Government. Distribution Statement "A" (Approved for Public Release, Distribution Unlimited)”.

\bibliography{Aibl}

@book{anderson2007can,
  author    = {Anderson, John R.},
  title     = {How Can the Human Mind Occur in the Physical Universe?},
  publisher = {Oxford University Press},
  address   = {New York},
  year      = {2007}
}

@article{blundell2016model,
  author  = {Blundell, Charles and Uria, Benigno and Pritzel, Alexander and Li, Yazhe and Ruderman, Avraham and Leibo, Joel Z. and Rae, Jack and Wierstra, Daan and Hassabis, Demis},
  title   = {Model-Free Episodic Control},
  journal = {arXiv preprint arXiv:1606.04460},
  year    = {2016}
}

@article{breunig2000lof,
  author  = {Breunig, Markus M. and Kriegel, Hans-Peter and Ng, Raymond T. and Sander, J{\"o}rg},
  title   = {{LOF}: Identifying Density-Based Local Outliers},
  journal = {Proceedings of the 2000 ACM SIGMOD International Conference on Management of Data},
  pages   = {93--104},
  year    = {2000}
}

@article{chandola2009anomaly,
  author  = {Chandola, Varun and Banerjee, Arindam and Kumar, Vipin},
  title   = {Anomaly Detection: A Survey},
  journal = {ACM Computing Surveys},
  volume  = {41},
  number  = {3},
  pages   = {1--58},
  year    = {2009}
}

@inproceedings{chapelle2011empirical,
  author    = {Chapelle, Olivier and Li, Lihong},
  title     = {An Empirical Evaluation of {T}hompson Sampling},
  booktitle = {Advances in Neural Information Processing Systems},
  volume    = {24},
  pages     = {2249--2257},
  year      = {2011}
}

@article{dal2017credit,
  author  = {Dal Pozzolo, Andrea and Le Borgne, Yann-A{\"e}l and Bontempi, Gianluca and Caelen, Olivier and Waterschoot, Serge},
  title   = {Learned Lessons in Credit Card Fraud Detection from a Practitioner Perspective},
  journal = {Expert Systems with Applications},
  volume  = {41},
  number  = {10},
  pages   = {4915--4928},
  year    = {2014}
}

@inproceedings{devlin2019bert,
  author    = {Devlin, Jacob and Chang, Ming-Wei and Lee, Kenton and Toutanova, Kristina},
  title     = {{BERT}: Pre-training of Deep Bidirectional Transformers for Language Understanding},
  booktitle = {Proceedings of the 2019 Conference of the North American Chapter of the Association for Computational Linguistics},
  pages     = {4171--4186},
  year      = {2019}
}

@misc{dua2019uci,
  author       = {Dua, Dheeru and Graff, Casey},
  title        = {{UCI} Machine Learning Repository},
  howpublished = {University of California, Irvine, School of Information and Computer Sciences},
  url          = {https://archive.ics.uci.edu/ml},
  year         = {2019}
}

@article{dutt2013cyber,
  author  = {Dutt, Varun and Ahn, Young-Suk and Gonzalez, Cleotilde},
  title   = {Cyber Situation Awareness: Modeling Detection of Cyber Attacks with Instance-Based Learning Theory},
  journal = {Human Factors},
  volume  = {55},
  number  = {3},
  pages   = {605--618},
  year    = {2013}
}

@article{gershman2017reinforcement,
  author  = {Gershman, Samuel J. and Daw, Nathaniel D.},
  title   = {Reinforcement Learning and Episodic Memory in Humans and Animals: An Integrative Framework},
  journal = {Annual Review of Psychology},
  volume  = {68},
  pages   = {101--128},
  year    = {2017}
}

@article{gonzalez2003instance,
  author  = {Gonzalez, Cleotilde and Lerch, F. Javier and Lebiere, Christian},
  title   = {Instance-Based Learning in Dynamic Decision Making},
  journal = {Cognitive Science},
  volume  = {27},
  number  = {4},
  pages   = {591--635},
  year    = {2003}
}

@article{gonzalez2011instance,
  author  = {Gonzalez, Cleotilde and Dutt, Varun},
  title   = {Instance-Based Learning: Integrating Sampling and Repeated Decisions from Experience},
  journal = {Psychological Review},
  volume  = {118},
  number  = {4},
  pages   = {523--551},
  year    = {2011}
}

@article{graves2014neural,
  author  = {Graves, Alex and Wayne, Greg and Danihelka, Ivo},
  title   = {Neural {T}uring Machines},
  journal = {arXiv preprint arXiv:1410.5401},
  year    = {2014}
}

@article{harper2015movielens,
  author  = {Harper, F. Maxwell and Konstan, Joseph A.},
  title   = {The {MovieLens} Datasets: History and Context},
  journal = {ACM Transactions on Interactive Intelligent Systems},
  volume  = {5},
  number  = {4},
  pages   = {Article 19},
  doi     = {10.1145/2827872},
  year    = {2015}
}

@techreport{harris1999splice,
  author      = {Harries, Michael},
  title       = {{SPLICE}-2 Comparative Evaluation: Electricity Pricing},
  institution = {University of New South Wales, School of Computer Science and Engineering},
  year        = {1999}
}

@inproceedings{khandelwal2020generalization,
  author    = {Khandelwal, Urvashi and Levy, Omer and Jurafsky, Dan and Zettlemoyer, Luke and Lewis, Mike},
  title     = {Generalization through Memorization: Nearest Neighbor Language Models},
  booktitle = {International Conference on Learning Representations},
  year      = {2020}
}

@article{kolodner1992introduction,
  author  = {Kolodner, Janet L.},
  title   = {An Introduction to Case-Based Reasoning},
  journal = {Artificial Intelligence Review},
  volume  = {6},
  number  = {1},
  pages   = {3--34},
  year    = {1992}
}

@article{lebiere2025cognitive,
  author  = {Lebiere, Christian and Pirolli, Peter and Johnson, Matthew and Martin, Michael and Morrison, Don},
  title   = {Cognitive Models for Machine Theory of Mind},
  journal = {Topics in Cognitive Science},
  volume  = {17},
  number  = {2},
  pages   = {268--290},
  year    = {2025}
}

@inproceedings{lewis2020retrieval,
  author    = {Lewis, Patrick and Perez, Ethan and Piktus, Aleksandra and Petroni, Fabio and Karpukhin, Vladimir and Goyal, Naman and K{\"u}ttler, Heinrich and Lewis, Mike and Yih, Wen-tau and Rockt{\"a}schel, Tim and Riedel, Sebastian and Kiela, Douwe},
  title     = {Retrieval-Augmented Generation for Knowledge-Intensive {NLP} Tasks},
  booktitle = {Advances in Neural Information Processing Systems},
  volume    = {33},
  year      = {2020}
}

@inproceedings{li2010contextual,
  author    = {Li, Lihong and Chu, Wei and Langford, John and Schapire, Robert E.},
  title     = {A Contextual-Bandit Approach to Personalized News Article Recommendation},
  booktitle = {Proceedings of the 19th International Conference on World Wide Web},
  pages     = {661--670},
  year      = {2010}
}

@inproceedings{liu2008isolation,
  author    = {Liu, Fei Tony and Ting, Kai Ming and Zhou, Zhi-Hua},
  title     = {Isolation Forest},
  booktitle = {Proceedings of the 2008 Eighth IEEE International Conference on Data Mining},
  pages     = {413--422},
  year      = {2008}
}

@article{nyhus2009wisconsin,
  author  = {Nyhus, Erika and Barcel{\'o}, Francisco},
  title   = {The Wisconsin Card Sorting Test and the Cognitive Assessment of Prefrontal Executive Functions: A Critical Update},
  journal = {Brain and Cognition},
  volume  = {71},
  number  = {3},
  pages   = {437--451},
  year    = {2009}
}

@inproceedings{pritzel2017neural,
  author    = {Pritzel, Alexander and Uria, Benigno and Srinivasan, Sriram and Puigdom{\`e}nech, Adri{\`a} and Vinyals, Oriol and Hassabis, Demis and Wierstra, Daan and Blundell, Charles},
  title     = {Neural Episodic Control},
  booktitle = {International Conference on Machine Learning},
  pages     = {2827--2836},
  year      = {2017}
}

@inproceedings{reimers2019sentence,
  author    = {Reimers, Nils and Gurevych, Iryna},
  title     = {Sentence-{BERT}: Sentence Embeddings Using {S}iamese {BERT}-Networks},
  booktitle = {Proceedings of the 2019 Conference on Empirical Methods in Natural Language Processing},
  pages     = {3982--3992},
  year      = {2019}
}

@inproceedings{sakurada2014anomaly,
  author    = {Sakurada, Mayu and Yairi, Takehisa},
  title     = {Anomaly Detection Using Autoencoders with Nonlinear Dimensionality Reduction},
  booktitle = {Proceedings of the MLSDA 2014 2nd Workshop on Machine Learning for Sensory Data Analysis},
  pages     = {4--11},
  year      = {2014}
}

@article{scholkopf2001estimating,
  author  = {Sch{\"o}lkopf, Bernhard and Platt, John C. and Shawe-Taylor, John and Smola, Alex J. and Williamson, Robert C.},
  title   = {Estimating the Support of a High-Dimensional Distribution},
  journal = {Neural Computation},
  volume  = {13},
  number  = {7},
  pages   = {1443--1471},
  year    = {2001}
}

@inproceedings{snell2017prototypical,
  author    = {Snell, Jake and Swersky, Kevin and Zemel, Richard S.},
  title     = {Prototypical Networks for Few-Shot Learning},
  booktitle = {Advances in Neural Information Processing Systems},
  volume    = {30},
  year      = {2017}
}

@article{stocco2024integrated,
  author  = {Stocco, Andrea and Rice, Peter and Thomson, Robert and Smith, Benjamin and Morrison, Don and Lebiere, Christian},
  title   = {An Integrated Computational Framework for the Neurobiology of Memory Based on the {ACT-R} Declarative Memory System},
  journal = {Computational Brain \& Behavior},
  volume  = {7},
  number  = {1},
  pages   = {129--149},
  year    = {2024}
}

@inproceedings{stocco2026metacognition,
  author    = {Stocco, Andrea and Bushnell, Linda and Poovendran, Radha},
  title     = {Algorithmic Metacognition through Fast Inverse Instance-Based Learning},
  booktitle = {Proceedings of the 23rd International Conference on Cognitive Modeling},
  note      = {In press},
  year      = {2026}
}

@article{thompson1933likelihood,
  author  = {Thompson, William R.},
  title   = {On the Likelihood that One Unknown Probability Exceeds Another in View of the Evidence of Two Samples},
  journal = {Biometrika},
  volume  = {25},
  number  = {3/4},
  pages   = {285--294},
  year    = {1933}
}

@article{welford1962note,
  author  = {Welford, B. P.},
  title   = {Note on a Method for Calculating Corrected Sums of Squares and Products},
  journal = {Technometrics},
  volume  = {4},
  number  = {3},
  pages   = {419--420},
  year    = {1962}
}

@inproceedings{weston2015memory,
  author    = {Weston, Jason and Chopra, Sumit and Bordes, Antoine},
  title     = {Memory Networks},
  booktitle = {Proceedings of the International Conference on Learning Representations},
  year      = {2015}
}

@article{xu2022modeling,
  author  = {Xu, Tianxin and Singh, Karambir and Rajivan, Prashanth},
  title   = {Modeling Phishing Decision Using Instance-Based Learning and Natural Language Processing},
  journal = {Proceedings of the 55th Hawaii International Conference on System Sciences},
  pages   = {1--10},
  year    = {2022}
}

@article{yang2024reliance,
  author  = {Yang, Y. C. and Sibert, Catherine and Stocco, Andrea},
  title   = {Reliance on Episodic vs. Procedural Systems in Decision-Making Depends on Individual Differences in Their Relative Neural Efficiency},
  journal = {Computational Brain \& Behavior},
  volume  = {7},
  number  = {3},
  pages   = {420--436},
  year    = {2024}
}

@article{mitsopoulos2025cognitively,
  title={Cognitively-plausible reinforcement learning in epidemiological agent-based simulations},
  author={Mitsopoulos, Konstantinos and Baker, Lawrence and Lebiere, Christian and Pirolli, Peter and Orr, Mark and Vardavas, Raffaele},
  journal={Frontiers in epidemiology},
  volume={5},
  pages={1563731},
  year={2025},
  publisher={Frontiers Media SA}
}
\end{document}